# Analysis of Information Propagation in Ethereum Network Using Combined Graph Attention Network and Reinforcement Learning to Optimize Network Efficiency and Scalability

*Stefan Kambiz Behfar*[1,2], *Jon Crowcroft*[2,3]

[1]Department of Information Systems, Geneva School of Business Administration, Geneva, Switzerland
[2]Department of Computer Science and Technology, Cambridge University, Cambridge, UK
[3]Alan Turing Institute, London, UK

**Abstract**

Blockchain technology has revolutionized the way information is propagated in decentralized networks. Ethereum, as a major blockchain platform, plays a pivotal role in facilitating smart contracts and decentralized applications. Understanding information propagation dynamics in Ethereum is crucial for ensuring network efficiency, security, and scalability. In this study, we propose an innovative approach that utilizes Graph Convolutional Networks (GCNs) to analyze the information propagation patterns in the Ethereum network. The first phase of our research involves data collection from the Ethereum blockchain, consisting of blocks, transactions, and node degrees. We construct a transaction graph representation of the Ethereum network using adjacency matrices to capture the node embeddings; while our major contribution is to develop a combined Graph Attention Network (GAT) and Reinforcement Learning (RL) model to optimize the network efficiency and scalability. It learns the best actions to take in various network states, ultimately leading to improved Ethereum network efficiency and throughput and optimize gas limits for block processing. In the experimental evaluation, we analyze the performance of our model on a large-scale Ethereum dataset. We investigate effectively aggregating information from neighboring nodes capturing graph structure and updating node embeddings using GCN with the objective of transaction pattern prediction, accounting for varying network loads and number of blocks. Not only we design a gas limit optimization model and provides the algorithm, but also to address scalability, we demonstrate the use and implementation of sparse matrices in GraphConv, GraphSAGE, and GAT. The results indicate that our designed GAT-RL model achieves superior results compared to other GCN models in terms of performance. It effectively propagates information across the Ethereum network, optimizing gas limits for block processing and improving network efficiency.

## Introduction

The advent of blockchain technology has revolutionized the world of cryptocurrencies, with Ethereum being one of the most prominent platforms. Ethereum's decentralized nature and smart contract functionality have resulted in a massive amount of transactional data generated daily. As a result, efficient pattern recognition methods are essential for extracting meaningful insights and addressing security issues within the network.

A graph convolution algorithm based on deep learning techniques has been proposed to automatically generate features using a graph algorithm. Kipf and Welling (2017) and Li et al. (2018) used Graph Convolutional Network (GCN) model and its derivatives that apply semi-supervised classification using neural network models to graph data. Chen et al. (2020) suggested a GCN-based phishing detection model which samples subgraphs by random walk and applies node embeddings and a model to incorporate spatial structures and node features. Lin et al. (2020) investigated modeling the Ethereum transaction network as a weighted temporal graph and a Temporal Weighted Multidigraph Embedding to incorporate temporal and weighted transaction edges.
The basic homogeneous GNN models such as original GCN (Kipf and Welling, 2017) cannot be applied to large graph structures because of its computational cost to learn network embeddings. The GraphSAGE model (Hamilton et al, 2017) learns the aggregation function from neighboring nodes instead of embedding nodes. The significance of GraphSAGE is that the learned aggregation function is applicable to unknown nodes and reduces the computational functions for large graphs with neighboring node samplings. Furthermore, the GAT (graph attention network) model (Velickovic et al., 2017) implicitly learns the different weight of each neighborhood as so-called "attention".

In Ethereum transaction networks, some nodes have different attributes such as exchanges and wallets. Such graphs with different vertices and edges are called heterogeneous graphs, which is more informative for machine learning than a homogeneous graph with the usual types. GNN models corresponding to many heterogeneous graphs have been proposed because of their increasing application to knowledge graphs (Kanezashi et al, 2022). At the same time, we aim to use a combined GCN and RL model to learn the best actions to take in various network states, ultimately lead to improved Ethereum network efficiency and throughput.

While some approaches for smart contract vulnerability detection had been suggested using test methods from the programming language community (Luu et al., 2016; Tsankov et al., 2018), Zhuang (2020) proposed an approach using graph neural networks (GNNs) for smart contract vulnerability detection. Although GNNs have been extensively studied and applied in various domains, their application in blockchain networks, particularly for enabling information propagation by aggregating node representations has not been thoroughly explored. The literature gap is as follows:

- Sparse Literature on Information Propagation in Blockchain: While there are some studies on analyzing transactional data and extracting patterns from blockchain networks, there is limited research on utilizing GCN layers to facilitate information propagation in these networks. This includes aggregating information from neighboring nodes, capturing graph structure, and updating node embeddings in blockchain-specific context.
- Scalability and Efficiency Concerns: Applying GCNs to large-scale blockchain networks like Ethereum presents scalability and efficiency challenges. Blockchain networks have vast amounts of data, making information propagation and node representation updates computationally expensive. Research is needed to develop optimized and scalable GCN architectures suitable for blockchain networks.
- Smart Contract Interaction Modeling: There is a lack of studies focusing on modeling smart contract interactions as graph structures and leveraging GCNs to analyze and understand these interactions. It is challenging to accurately attribute the performance load of specific smart contract operations to the edges and nodes of the graph. This is because transactions involving smart contracts often have more complex interactions than simple value transfers between user addresses.

In this study, we focus on the 1st and 2nd research gap in utilizing graph convolutional layers to facilitate information propagation and leave the 3rd to our next paper. We, therefore, investigate effectively aggregating information from neighboring nodes, capturing and updating graph structure. Initially, we discuss the model design, and explanation of how multi-layer GNN works, and provide a novel approach to use GAT and RL to optimize network efficiency. In the empirical analysis, we download the Ethereum network for a specific block range and apply GCN to update node embedding. We design an algorithm for Gas limit optimization, apply GCN to a large-scale Ethereum network, do the comparison between different sparse matrix method in GraphConv, GraphSAGE, and GAT models, and show how GAT-RL renders a better performance.

## Model Design

This section presents an overview of Convolutional Neural Networks (CNNs) and their suitability for pattern recognition tasks. CNNs have achieved state-of-the-art performance in various image recognition problems due to their ability to automatically learn hierarchical features from raw data. Researchers have recognized the potential of CNNs in tackling complex blockchain data analysis, particularly in Ethereum. GCNs are a type of neural network architecture designed for graph-structured data. They utilize graph convolutional layers (GCLs) to aggregate information from neighboring nodes and edges to update node embeddings. Here's a step-by-step explanation of how GCLs work:

*Graph Representation:*
A graph consists of nodes and edges, where nodes represent entities, and edges represent relationships between those entities. In the context of applying GCNs to the graph, each node is associated with a feature vector that encodes the information related to that node.

*Adjacency Matrix:*
The adjacency matrix represents the connections between nodes in the graph. It is a binary matrix where each entry *A[i, j]* is 1 if there is an edge between nodes i and j, and 0 otherwise. For directed graphs, *A[i, j] = 1* indicates a directed edge from node i to node j.

*Node Features:*
Each node in the graph is associated with a feature vector. These feature vectors can represent attributes, properties, or characteristics of the nodes.

*Graph Convolutional Layer:*
The graph convolutional layer aggregates information from neighboring nodes and edges to update the node embeddings. The core operation is the graph convolution, which can be defined as follows:

$$h_i^{(l+1)} = \sigma\left(\sum_{N(v_i)} W^{(l)} h_i^{(l)} + b^{(l)}\right)$$

Where,
$\mathcal{N}(v)$ denotes the set of neighboring nodes of node v in the graph.
$h_i^{(l)}$ represents the node embedding of node i in the l-th layer.
$W^{(l)}$ is the weight matrix of the l-th layer.
$b^{(l)}$ is the bias term of the l-th layer.
σ is the activation function (e.g., ReLU or Sigmoid) that introduces non-linearity.
Σ is the summation over neighboring nodes of node i.

*Aggregation of Information:*
In the graph convolutional layer, each node aggregates information from its neighbors using the adjacency matrix. The aggregation process captures the information from neighboring nodes and edges, allowing the node to learn from its local neighborhood.

*Node Embedding Update:*
After the aggregation step, the graph convolutional layer updates the node embeddings based on the aggregated information. This update captures the local and global graph structure and can help improve the node representations for downstream tasks like node classification or link prediction.

*Stacking Multiple Layers:*
GCNs can be stacked to create deeper architectures, allowing the model to learn more complex representations of the graph data. Each layer receives the updated node embeddings from the previous layer and performs another round of aggregation and node embedding updates.

By applying graph convolutional layers, the model can effectively aggregate information from neighboring nodes and edges, enabling it to learn meaningful representations of the graph structure and perform tasks such as node classification, link prediction, or graph-level regression.

## Graph Convolution Operation

Deriving a complete mathematical proof of Graph Neural Networks (GNNs) can be quite involved and requires a deep understanding of linear algebra, graph theory, and optimization. Here, we intend to give an overview of the fundamental concepts for less-familiar readers to gain background necessary:

1. Basic Definitions:

Graph Representation: A graph is represented as $G = (V, E)$, where $V$ is the set of nodes (vertices) and E is the set of edges connecting the nodes.
Node Features: Each node $v_i$ in the graph is associated with a feature vector $x_i$, representing its attributes or characteristics.
Aggregation: Aggregation is a process to combine the feature vectors of neighboring nodes to obtain a summary representation.

2. Node Aggregation:

The aggregation operation is typically defined as a weighted sum of the feature vectors of neighboring nodes, using an adjacency matrix $A$ to capture graph connectivity:

$$h_i = Aggregate(x_i | j \in N(i)) = \sum_{j \in N(i)} A_{ij} x_j$$

Where $h_i$ is the aggregated representation of node $i$, $N(i)$ is the set of neighbors of node $i$, and $A_{ij}$ represents the weight between node $i$ and node $j$.

3. Weight Sharing and Parameterization:

GNNs leverage weight sharing across nodes, which allows the same parameters to be used for aggregation and transformation of each node in the graph. This weight sharing enables GNNs to generalize across different nodes and learn graph-level patterns. Parameterization involves learning the weight matrix $W$ that will be shared across all nodes. The aggregated features are then transformed using the shared weight matrix:

$$h_i = Activation\left(W \, Aggregate\left(x_j | j \in N(i)\right)\right)$$

4. Multi-Layer GNN:

To enable information propagation across multiple layers, the graph convolution operation is performed iteratively through multiple graph convolutional layers. The output of one layer serves as the input to the next layer, allowing the propagation of information through the network. The node representations are updated iteratively layer by layer, allowing information from neighbors and their neighbors to be incorporated into the node features. By applying these mathematical formulas, Graph Convolutional Layers enable information propagation in the blockchain network by aggregating and updating node representations based on the neighboring nodes and graph structure. This process facilitates the learning of meaningful representations that capture the dependencies and patterns in the blockchain network. In summary, a multi-layer GCN applies aggregation and transformation operations iteratively to propagate information across the graph, capturing complex graph patterns and relationships. The parameters $W^l$ are learned during the training process to optimize the model's performance on a specific graph-based task. GNNs often consist of multiple layers, where each layer iteratively updates the node representations:

$$h_i^{(l)} = Activation\left(W^{(l)} Aggregate\left(h_j^{(l-1)} | j \in N(i)\right)\right)$$

Here, $h_i^{(l)}$ is the representation of node $i$ at layer $l$, and $h_j^{(l-1)}$ is the representation of neighboring node $j$ at the previous layer $(l-1)$. The formula for a multi-layer graph convolutional operation in a Graph Convolutional Network is based on aggregating information from neighboring nodes and applying multiple layers of transformations.

5. Output Layer and Optimization:

For specific tasks like node classification, the final layer of the GNN is usually followed by a global pooling operation to obtain the graph-level representation. The pooled representation is then used to make predictions. The GNN parameters (weight matrices) are learned through optimization using gradient-based methods, such as stochastic gradient descent (SGD) or Adam.

## GAT and RL to optimize network efficiency

To apply mathematical formulas to analyze the Ethereum network using a combination of Graph Attention Network (GAT) and Reinforcement Learning (RL) to optimize network efficiency, we will outline the key mathematical concepts involved in each component:

1. Graph Attention Network (GAT)

- Attention Coefficients: In GAT, attention coefficients ($\alpha_{ij}$) are computed to capture the importance of neighboring

nodes in aggregating node features. These coefficients are calculated using an attention mechanism as follows:

$$e_{ij} = LeakyReLU(a^T [W * h_i || W * h_j])$$

$$\alpha_{ij} = softmax(e_{ij}) = exp(e_{ij}) / \sum_k exp(e_{ik})$$

Where $h_i$ and $h_j$ are the feature representations of nodes i and j, W is a learnable weight matrix, and a is a shared parameter.
- Aggregation with Attention: After computing the attention coefficients, the features of node i ($h_i$) are aggregated based on their neighbors' importance using the attention coefficients ($\alpha_{ij}$):

$$h'_i = \sum_j \alpha_{ij} * (W * h_j)$$

The aggregated feature $h'_i$ is then used for subsequent computations.

2. Reinforcement Learning (RL)

- Q-Value Function: In RL, the Q-value function Q(s, a) represents the expected total reward obtained by taking action 'a' in state 's' and following the optimal policy thereafter. In DQN, the Q-value function is approximated using a neural network.

$$Q(s, a) = f(s, a; \theta)$$

Where 's' is the state, 'a' is the action, and θ are the parameters of the Q-network.
- Bellman Equation: The Bellman equation defines the optimal Q-value function as the maximum expected return achievable by taking best action in a given state.

$$Q^*(s, a) = E[R_{t+1} + \gamma * max_{a'} Q^*(s', a')]$$

Where $R_{t+1}$ is the reward received at the next time step, γ is the discount factor, and s' is the next state.
- Q-Learning Update: The Q-learning update rule adjusts the Q-values based on the reward and the optimal action in the next state.

$$\Delta Q(s, a) = \alpha * (r + \gamma * max_{a'} Q(s', a') - Q(s, a))$$
$$Q(s, a) \leftarrow Q(s, a) + \Delta Q(s, a) \alpha_{ij}$$

Where α is the learning rate and 'r' is the reward obtained by taking action 'a' in state 's'.

3. Optimization for Network Efficiency

- RL Agent's Policy: In the context of optimizing Ethereum network efficiency, the RL agent's policy defines how it selects actions (e.g., adjusting gas limits) to minimize block processing time and maximize throughput. The RL agent explores and exploits the environment using an exploration-exploitation strategy. For example, it can use ε-greedy exploration, where with probability ε, the agent selects a random action, and with probability (1-ε), it selects the action with the highest Q-value. The goal of the RL agent is to find the optimal policy π* that maximizes the expected cumulative reward over time:

$$\pi^* = argmax_\pi E_\pi [\Sigma_t \gamma^t * r_t]$$

Where:
$E_\pi$ denotes the expectation under policy π.
$\Sigma_t$ represents the summation over all time steps t.
$\gamma^t$ is the discount factor raised to the power of t.
- Reward Function: The reward function provides feedback to the RL agent about the efficiency of the selected action. For Ethereum network analysis, it can be formulated as:

$$reward = -block\_processing\_time$$

Where 'block_processing_time' represents the time taken to process a block with the chosen gas limit.
- Training Objective: The RL agent aims to find the optimal policy that maximizes the expected total reward over time by interacting with the Ethereum network. It interacts with the Ethereum network, observes states (e.g., current block data, pending transactions, network congestion), and updates its policy based on the received rewards. The agent learns the best actions to take in various network states, ultimately leading to improved Ethereum network efficiency and throughput.

Let's consider a simplified Ethereum network with a fixed set of transactions and a single miner. The objective is to find the optimal gas limit that maximizes throughput, assuming each transaction takes a fixed time to process and gas usage is linearly related to time. Let:

N be the total number of transactions in the block.
G be the gas limit (i.e., the maximum amount of gas that can be used in the block).
T be the time taken to process a single transaction (assumed fixed).

The total time taken to process all transactions in the block can be represented as:

$$Total\ processing\ time = N * T$$

To maximize throughput, we want to maximize the number of transactions processed in the block while respecting the gas limit. We represent the total gas used in the block as:

$$Total\ gas\ used = N * G$$

If the total gas used (N * G) exceeds the gas limit, then the block will be full, and the number of transactions processed will be limited by the gas limit. Therefore, the number of transactions processed in the block can be represented as:

$$Transactions\ processed = min(N, G / T)$$

To maximize throughput, we need to find the gas limit G that maximizes the number of transactions processed:

$$\text{maximize Transactions processed}$$
$$= min(N, G / T) \quad [subject\ to\ G >= 0]$$

Since the number of transactions processed is a linear function of G (piecewise linear due to the "min" operation), we can differentiate it with respect to G and set the derivative to zero to find the maximum:

$$d/dG\ [min(N, G / T)] = 0$$

To handle the "min" operation, we consider two cases:
Case 1: When G / T > N, the maximum number of transactions processed will be N.

$$d/dG\ [N] = 0$$

Case 2: When G / T <= N, the maximum number of transactions processed will be G / T.

$$d/dG\ [G / T] = 0$$

Solving for G in Case 2: 1 / T = 0
Since the derivative is zero in Case 2, we cannot find a valid solution for G that maximizes the number of transactions processed. Therefore, the maximum throughput is achieved when G / T > N. In conclusion, in a simplified Ethereum network where each transaction takes a fixed time to process and gas usage is linearly related to time, the optimal gas limit that maximizes throughput is when **G / T > N**, where N is the total number of transactions in the block. This means that the gas limit should be set high enough to allow the processing of all transactions in the block without reaching the gas limit constraint.

## Empirical Analysis

### Datasets and Data Preprocessing
This section discusses the publicly available Ethereum datasets and how we obtain them. Additionally, data preprocessing steps used to prepare data for CNN-based analysis will be highlighted. Creating a complete transaction graph for all Ethereum blocks would be a computationally intensive task, as it would involve processing and storing a large amount of data. Additionally, the Ethereum blockchain is constantly growing, so the graph would continuously expand as new blocks are added. However, we provide an algorithm to generate a transaction history graph for a range of blocks. The following simple algorithm demonstrate how to create a graph of transactions between Ethereum addresses within a specified block range.

1. Place <your_ethereum _node_url> with the URL of the Ethereum node obtained from Infura/Alchemy website in order to access to Ethereum Mainnet.
2. Set the start_block and end_block variables to specify the block range to create the transaction history graph; one can create a descending *range(latest_block, 0, -1)*.
3. The create_transaction_history_graph function.

### Optimal Gas Limit Design
Algorithm1 presents an algorithm designed to determine optimal gas limit that maximize Ethereum network efficiency.

Algorithm 1: designed to determine optimal gas limits

1. Initialization:
   - Initialize the gas limit (**gas_limit**) to a starting value.
   - Set the gas limit increment (**gas_limit_increment**) to fine-tune adjustments.
   - Define the maximum allowed gas limit (**max_gas_limit**) to prevent excessive values.
   - Specify the target block processing time (**target_time**) for efficient block inclusion.
   - Set the congestion threshold (**congestion_threshold**) to avoid network congestion.
2. Iterative Process:
   a. Simulate Block Processing:
   - Calculate the expected block processing time based on the average transaction processing time and the number of transactions included in the block.
   - Determine the congestion level based on the gas limit and the number of transactions with higher gas fees.
   b. Check Convergence:
   - If the expected block processing time is close to the target time and the congestion level is below the threshold, the algorithm terminates.
   c. Adjust Gas Limit:
   - If the expected block processing time is too high or the congestion level is too high, reduce the **gas_limit** by **gas_limit_increment**.
   - If the expected block processing time is too low or the congestion level is too low, increase the **gas_limit** by **gas_limit_increment**.
3. Algorithm Convergence:
   - The algorithm repeats the iterative process until it converges or reaches a predefined maximum number of iterations.

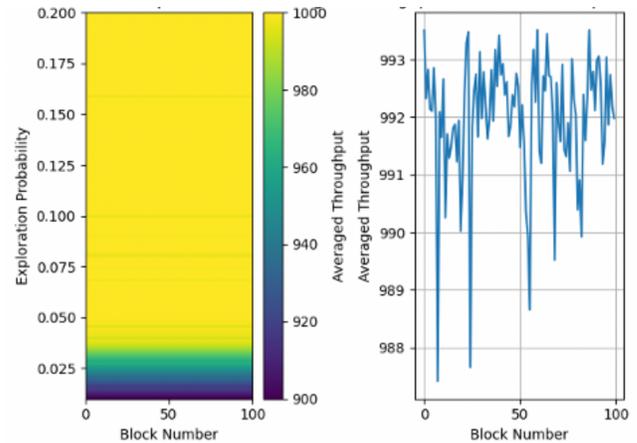

Figure 1: the average throughput achieved by the RL agent

The plotted results in Figure 1 show the learning progress of the reinforcement learning (RL) agent as it attempts to optimize gas limits for block processing in the Ethereum network. The x-axis represents the blocks, while the y-axis represents the throughput (number of transactions processed) for each block. Here we explain the results:

1. Line Plots:

The line plot represents the throughput achieved by the RL agent in different episodes and block number. The agent starts with random gas limits and gradually learns to adjust the gas limits to maximize throughput.

2. Learning Progress:

As the number of block number increases, the graph show how the agent's throughput improves over time. Initially, the agent explores various gas limit choices, leading to fluctuations in throughput. As shown, the agent learns, the throughput values stabilize and tend to increase.

3. Convergence:

The convergence of the RL agent can be observed when the lines start to converge to a more stable throughput pattern. This indicates that the agent has learned optimal gas limits for maximizing throughput.

4. Exploration-Exploitation Trade-off:

The lines might show some exploration and exploitation behavior. Initially, the agent explores different gas limit options (randomly selecting actions) to learn about their effects on throughput. As the agent gains knowledge, it starts exploiting its learned Q-values to choose better actions and converge to a higher throughput.

5. Final Throughput:

The randomness in the lines is due to the inherent stochasticity in the simulation of the environment and RL updates. The final throughput achieved after the RL agent's convergence is the most critical outcome of the learning process. A higher final throughput indicates that the RL agent has successfully optimized gas limits to maximize transaction processing.

In Figure 2, we illustrate the gas limit optimization achieved by the RL agent across all episodes and plot the average gas limit over blocks, learning rate and exploration probabilities. This provides a representation of the RL agent's learning progress and performance in maximizing the throughput. As the number of block increases, the optimized gas limit decreases, also this holds for exploration prob.

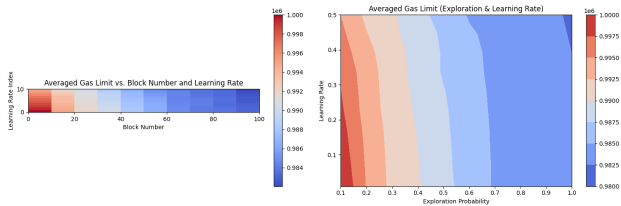

Figure 2: optimizing gas limit achieved by the RL agent

## Apply GCN to Large Scale Network

Applying GCNs to a large-scale Ethereum network to address scalability and efficiency challenges demands careful consideration of the data representation and the GCN architecture. Its implementation for a large-scale Ethereum network requires optimizations like using sparse matrices and distributed computing to handle the massive amount of data efficiently. Additionally, one might need to design custom GCN architectures to achieve better performance for Ethereum network analysis. To handle such scenarios, one common approach is to use GraphSAGE (Graph Sample and Aggregated) or Graph Attention Networks (GAT), which enable better scalability and capture more complex relationships in the data. We implement Sparse matrices using PyTorch Geometric for network scaling, then compare the three methods for different number of Ethereum blocks.

### Scaling GCNs Using Sparse Matrices

Scaling GCNs using sparse matrices involves modifying the conventional GCN equations to efficiently handle sparse adjacency matrices. The mathematical formulation can be described as follows:

1. Graph Convolutional Layer (GraphSAGEConv)

The GraphSAGEConv layer performs aggregation over the neighbors of each node using the GraphSAGE approach, which involves sampling a fixed-size neighborhood for each node and aggregating the features of the sampled neighbors. Given an input feature matrix X of shape (N, D), where N is the number of nodes, D is the number of input features per node, and the sparse adjacency matrix A of shape (N, N), the output feature matrix H at each GraphSAGEConv layer is calculated as:

$$H = \sigma(A @ X @ W)$$

Where, @ represents matrix multiplication.
σ is the activation function, typically ReLU or another non-linear function.
W is the weight matrix of the layer, which needs to be learned during training.

2. GraphSAGE Neighbor Sampling:

To handle large graphs efficiently, GraphSAGE employs neighbor sampling, which samples a fixed number (K) of neighbors for each node. This helps reduce memory consumption during training and speeds up computation. The equation for neighbor sampling involves selecting a random subset of neighbors for each node i based on the adjacency matrix A:

$$N_i = sample\_neighbors(A, i, K)$$

sample_neighbors (A, i, K) is a function that returns a set of K sampled neighbors for node i from adjacency matrix $A$.

3. Aggregation:

The sampled neighbor nodes' features are aggregated using mean or sum pooling to create the feature representation for each node's neighborhood.

$$H_{N_i} = sum\_pooling(H[N_i]) / K$$

- $H[N_i]$) represents the feature matrix of the sampled neighbors $N_i$.
- sum_pooling and mean_pooling are functions that calculate the sum or mean of the rows of a matrix.

    4. GraphSAGEConv Update:

The output feature matrix $H$ is updated with the aggregated neighborhood features to create the final node representations for the next layer. By handling the adjacency matrix as a sparse tensor, we can save memory and computation time when dealing with large graphs. We use PyTorch's sparse_coo_tensor to create a sparse matrix, and measure performance and scalability. PyTorch's sparse_coo_tensor is used to create a sparse matrix in CSC format. This format represents a sparse matrix by storing only the non-zero elements and their corresponding row and column indices. Using sparse matrices can significantly reduce memory consumption and speed up computations when working with large graphs.

**Scaling GCNs using GAT-RL**

To scale Graph Convolutional Networks (GCNs) using sparse matrices with GAT layer, we'll utilize PyTorch Geometric and its sparse tensor capabilities. GAT introduces attention mechanisms to focus on relevant neighbors during aggregation, making it memory-efficient for large graphs. The attention mechanism in GAT is used to compute attention coefficients for each pair of nodes, indicating the importance of one node's features for another node's representation. The aggregated representation for each node is computed by taking the weighted sum of its neighboring nodes' features, where the weights are the attention coefficients. The aggregated node representations $h'_i$ are updated using skip-connection, bias b, and weight vector a:

$$h'_i = \sigma\left(\sum_j \alpha_{ij} * (W * h_j) + b\right) * a$$

We show Accuracy and Scalability/Performance between GraphConv, GraphSAGE, and GAT in Table 2, for different number of blocks. GAT appears to outperform others.

Table 2. Comparison of Accuracy/Performance for GCNs.

| Comparison Accuracy/ Performance (sec) | Scaling GCNs using GraphConv | Scaling GCNs using GraphSAGE | Scaling GCNs using GAT model |
|---|---|---|---|
| #blocks =1000 | 0.7300/ 0.8577 | 0.990/ 0.635 | 0.9751/ 1.2146 |
| #blocks =3000 | 0.9948/ 0.8524 | 0.998/ 0.558 | 0.7307/ 1.9887 |
| #blocks =9000 | 0.8124/17.289 | 0.297 /3.870 | 0.9341/ 24.467 |

Our combined GAT-RL model is defined with Proximal Policy Optimization (PPO) RL and sparse tensor support. The code algorithm given in Algorithm 2 trains the GAT-RL model on our Ethereum dataset and measures its performance after each epoch for 10 epochs, however we train GAT and GAT-RL (PPO) separately with num_epochs = 1000 in our model.

Algorithm 2: algorithm designed for a GAT-RL model.

```
# Define the blockchain graph data (adjacency matrix and node features)
# Create train and test masks for inductive learning
# Define GAT model
# Train GAT model:
for each epoch in range(num_epochs):
    Train GAT model on the training data
    Compute and store the GAT loss
# Define the Ethereum optimization RL environment:
class EthereumOptimizationRL:
    Initialize environment with node features, train, and test mask
    Define apply_resource_allocation method to modify node features
    Define calculate_reward method to compute normalized rewards
# Define PPO policy and value function networks
# Train PPO agent using PPO algorithm:
for each epoch in range(num_epochs):
    Reset environment
    while not done:
        Sample action using policy network
        Apply action to environment
        Calculate normalized reward using the environment
        Store state, action, log probability, value, and normalized reward
    Compute advantages using normalized rewards
    Update policy and value function using PPO loss
# Train GAT-RL in a combined manner:
for each epoch in range(num_epochs):
    Train GAT model
    Create Ethereum Optimization RL environment
    Train RL agent using PPO algorithm
    Store GAT loss and PPO loss
# Plot GAT and PPO losses
```

In Figure 3, we compare the performance (Loss) of GAT versus GAT-RL which apparently performs better.

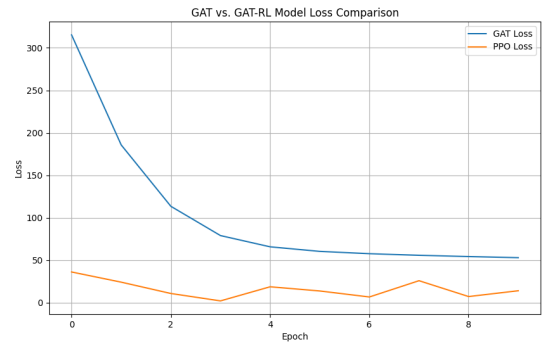

Figure 3: Loss comparison for GAT vs. GAT-RL models

**Conclusion**

By consolidating the existing knowledge from the literature, this study provides information propagation analysis of the Ethereum network using a combined GAT-RL where we focused on the research gap in utilizing graph convolutional layers to facilitate information propagation. Applying GCNs to large-scale blockchain networks like Ethereum presents scalability and efficiency challenges. Our major contribution was to develop a model to optimize network efficiency and show that it outperforms other GCN methods.